\documentclass[twocolumn]{cinc}
\usepackage{graphicx}

\begin{document}

\bibliographystyle{cinc}

\title{Temporal Training Strategies for Left Atrium and Left Atrial Appendage Segmentation in Dynamic Contrast 4DCT}


\author {David Montalvo-García$^{1,2}$, Lauren Severance$^{3}$, Elliot R. McVeigh$^{3,4,5}$, María J. Ledesma-Carbayo$^{1,2}$ \\
\ \\ 
{\fontsize{11}{14}\selectfont
$^1$ Biomedical Image Technologies, ETSI Telecomunicación, Universidad Politécnica de Madrid, Madrid, Spain \\
$^2$  Centro de Investigación Biomédica en Red de Bioingeniería, Biomateriales y Nanomedicina (CIBER-BBN), Instituto de Salud Carlos III, Madrid, Spain  \\
$^3$  Department of Bioengineering, University of California San Diego, La Jolla, CA, USA \\
$^4$  Division of Cardiovascular Medicine, University of California San Diego, La Jolla, CA, USA  \\
$^5$  Department of Radiology, University of California San Diego, La Jolla, CA, USA}
}

\maketitle

\begin{abstract}

Dynamic contrast-enhanced cardiac CT enables time-resolved analysis of contrast filling and washout in the left atrium (LA) and left atrial appendage (LAA), with potential applications for assessing blood stasis in atrial fibrillation (AF). Accurate segmentation across all frames is required for such analysis but is challenging due to large temporal contrast variations and the use of a single annotation per registered sequence. This creates a trade-off between training for robustness and limiting label noise. In this study, we investigate how temporal training-set design affects nnUNet-based segmentation of the LA and LAA in dynamic 4DCT. We compare training using a minimal two-frame dataset reflecting standard clinical practice, a physiologically selected subset of frames, and the full 27-frame sequence. We further evaluate the impact of foreground-based normalization. Training with all frames yielded the best performance in early low-contrast phases. However, the physiologically selected subset achieved comparable performance from the filling phase onward. Applying normalization parameters derived from the full dataset improved performance of reduced datasets in low-contrast frames, but did not fully close the gap. These findings highlight the importance of temporal diversity in training data for robust segmentation in dynamic CT, while indicating that carefully selected frame subsets may provide an effective trade-off between performance and efficiency for downstream applications.


%
%



\end{abstract}

\section{Introduction}

Atrial fibrillation (AF) is the most common arrhythmia, characterized by irregular atrial contractions, leading to slow blood flow in the left atrium (LA) and left atrial appendage (LAA), where thrombi can form and embolize, causing strokes \cite{palaniappan20262026}. The vast majority of thrombi in patients with AF originate in the LAA \cite{blackshear1996appendage}. Consequently, patient-specific evaluation of blood stasis in the LA and LAA may have substantial clinical value.

Dynamic contrast-enhanced cardiac CT has recently emerged as a promising technique for visualizing contrast filling and washout within heart chambers, enabling time-resolved analysis of atrial blood transport \cite{severance2024estimation}. A prerequisite for such analysis is accurate segmentation of the LA and LAA throughout the 4DCT sequence. This task is challenging because image contrast changes markedly over time, from early low-enhancement frames to peak-opacification and late washout phases.

Automatic segmentation is a key step for quantitative analysis of cardiac imaging. Deep learning methods have become the standard for this task, with convolutional neural networks achieving high accuracy across modalities. Among them, nnUNet \cite{isensee2021nnunet} is the state-of-the-art framework with a self-configuring design, which automatically adapts preprocessing, architecture, and training to the training dataset. However, applying it to dynamic imaging remains challenging because temporal contrast variations and limited annotations can significantly impact performance.

In this work, we investigate how temporal training-set design affects nnUNet segmentation performance in dynamic contrast 4DCT in patients with AF. We compare training with a minimal two-frame dataset available in routine clinical practice, a physiologically selected subset of frames, and the full sequence. We further assess whether part of the performance gap can be explained by nnUNet’s foreground-based CT normalization. Our goal is to identify an effective training strategy for robust LA and LAA segmentation across the different phases of the sequence, as a step toward downstream tasks such as automatic quantification of LAA blood transport.

\section{Methods}

\subsection{Study cohort and imaging protocol}
Patients undergoing evaluation for AF at UC San Diego Health were prospectively enrolled between March 2022 and October 2024 under institutional review board approval, and all participants provided informed consent. The study included 94 patients (mean age 74.4 ± 9 years; 29\% female). At the time of the scan, 22 patients were in sinus rhythm, 4 were atrial-paced, and 68 were in AF.

Gated coronary computed tomography angiography was performed using a 256-slice GE Healthcare (Chicago, IL) Revolution CT scanner. The time to peak contrast in the LA, measured using a timing bolus scan, was used to guide the scan timing.  The dynamic contrast 4DCT scan consisted of 27 CT volumes, each obtained at end systole following contrast injection. The acquisition was designed to sample contrast filling, peak enhancement, early washout, and delayed phases, including a delayed volume acquired approximately 60s after LA peak contrast.

\subsection{Image pre-processing and annotation}

Volumes were reconstructed with whole-heart coverage at approximately 0.5 mm in-plane resolution and 1.25 mm slice thickness using vendor deep-learning reconstruction algorithm \cite{hsieh2019new}. Respiratory motion was compensated using non-rigid registration \cite{thiruvenkadam2016gpnlperf}, and all frames in a sequence were aligned to a common reference space. Manual segmentations of the LA and LAA were created from the peak contrast frame image. This annotation was then propagated to all registered frames within the sequence and used as a common ground truth.

\subsection{Temporal training datasets}

To evaluate the effect of temporal sampling on segmentation performance, we defined three training sets:

\begin{itemize}
    \item Training Set \#1: frames 15 (peak contrast, based on acquisition protocol) and 27 (delayed phase), both of which are widely available in routine clinical practice.
    \item Training Set \#2: physiologically selected frames corresponding to the LAA input peak minus 5 seconds, the LAA input peak, the LAA top-voxel peak, and frame 27. The corresponding frame indices were determined offline for each subject from the annotated dynamic sequence and were used only to construct the training subsets; no such information is required at inference time.
    \item Training Set \#3: all available frames of each sequence. 
\end{itemize}

 Data were split into 56 training and 38 test subjects, with 5-fold cross-validation applied to the training set. All splits were performed at the subject level to prevent leakage of frames from the same sequence across training, validation and test sets. Stratification accounted for pacing artifacts, rhythm status during CT, and the delay between input and distal LAA contrast peaks, to ensure differences in LAA contrast filling were accounted for during training. 

\subsection{Model training}

All models were trained using the nnUNet \cite{isensee2021nnunet} framework with default settings and identical training procedures, varying only the temporal composition of the training data. Performance was evaluated on the same held-out test subjects across all available time frames using Dice and 95th-percentile Hausdorff distance (HD95) metrics.

Because nnUNet derives CT intensity normalization statistics from foreground voxels, the resulting parameters depend on the distribution of contrast phases in the training set. For CT images, nnU-Net applies a global foreground-based normalization scheme in which intensities are clipped using the 0.5th and 99.5th foreground percentiles and subsequently standardized using the global foreground mean and standard deviation computed over the training data. To assess the impact of this dependency, we performed a normalization ablation. Specifically, we computed the clipping and standardization parameters using Training Set \#3, which contains the full range of contrast enhancement, including low-intensity early-phase frames. The resulting statistics (foreground percentiles, mean, and standard deviation) were then used to normalize the images when training models on Training Sets \#1 and \#2. All normalization statistics were computed using foreground voxels from the training subjects only, without using any information from the held-out test set.

\section{Experiments and Results}

\subsection{Effect of temporal training set design}

\begin{figure}[h]
    \centering
    \includegraphics[width=0.98\linewidth]{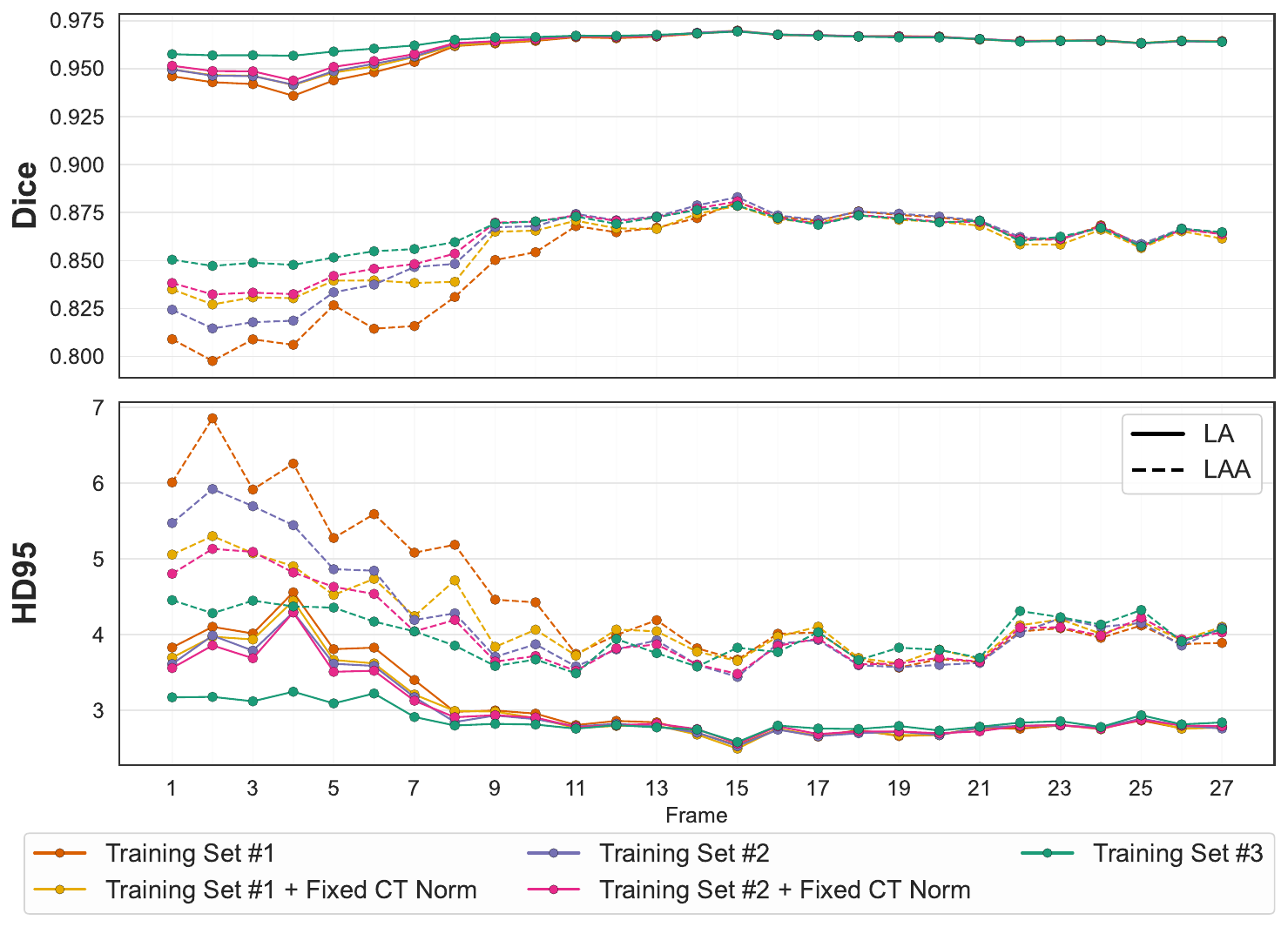}
    \caption{Segmentation performance across the dynamic sequence on test subjects. Mean Dice and HD95 for LA and LAA are shown as a function of frame number for models trained on all dataset, including variants using normalization params. derived from Training Set \#3.}
    \label{fig:curves}
\end{figure}

\begin{figure}[h]
    \centering
    \includegraphics[width=0.96\linewidth]{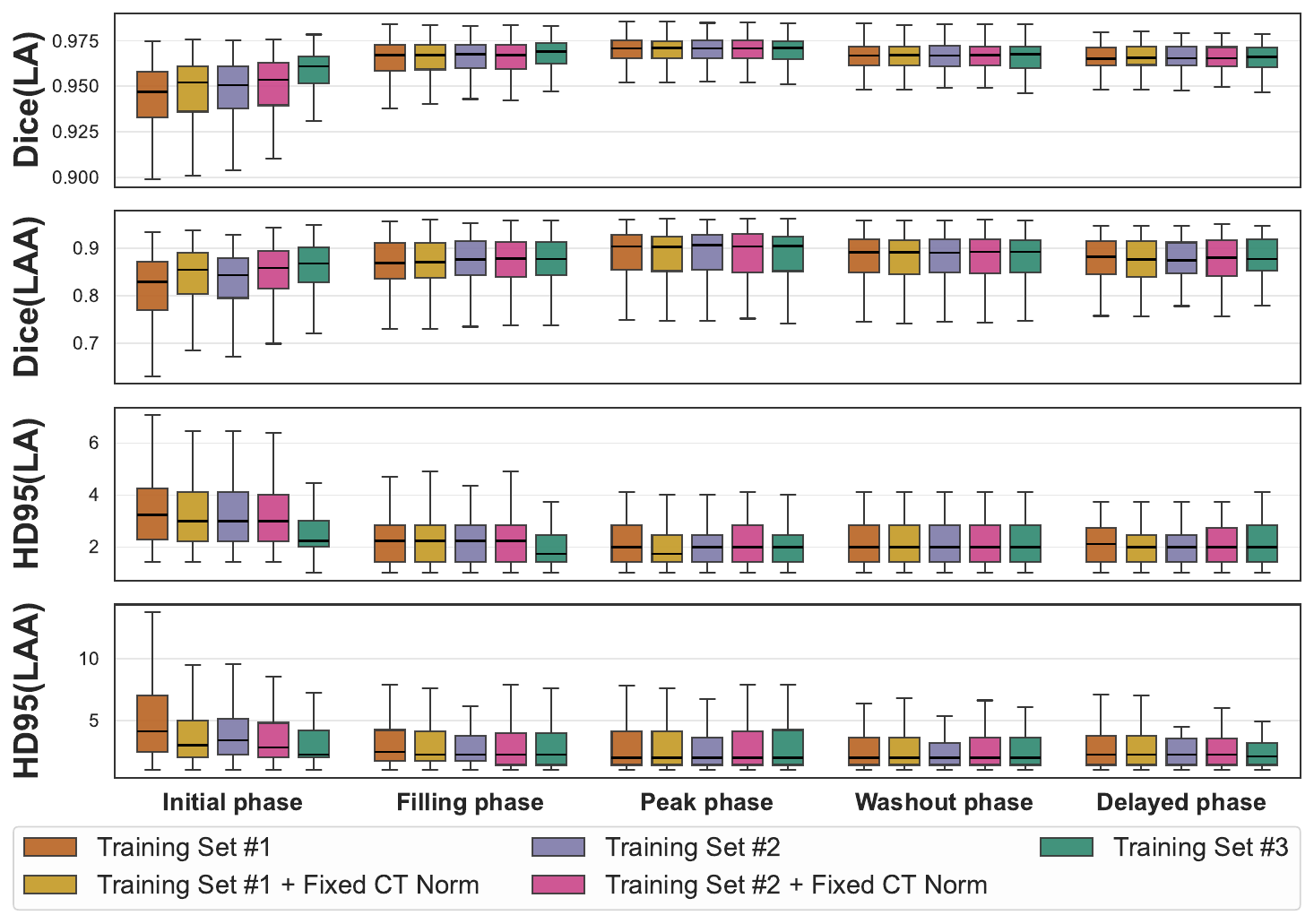}
    \caption{Phase-based segmentation performance on test subjects. Dice and HD95 are summarized as box plots across five phases (initial, filling, peak, washout, and delayed) for all models.}
    \vspace{-5px}
    \label{fig:boxplot}
    
\end{figure}

We first evaluate how the temporal composition of the training sets affects segmentation performance across the dynamic sequence. As shown in Fig.~\ref{fig:curves}, models trained with all frames (Training Set \#3) achieved the highest accuracy in early time frames, where contrast enhancement in the LA and LAA is low. However, models trained on reduced datasets (Training Sets \#1 and \#2) showed decreased performance in this initial phase. From the filling phase onward, the performance gap narrowed. Phase-based analysis (Fig.~\ref{fig:boxplot}) confirms that Training Set \#3 performs best in the initial phase, while all models exhibit similar Dice during peak and delayed phases.

Analysis of the HD95 metric in Fig.~\ref{fig:curves} reveals a complementary trend. While Training Set \#3 provides superior robustness in early frames, models trained on reduced sets achieve lower HD95 distances at peak enhancement (frame 15) and during the washout phase, particularly in the LAA. This suggests improved boundary delineation in high-contrast conditions when training is focused on representative frames.

Overall, these results indicate that exposure to low-contrast frames during training is critical for robust performance in early phases, whereas targeted selection of high-contrast frames can yield improved boundary accuracy in specific phases. This underscores the importance of aligning temporal dataset design with the intended application.

\subsection{Effect of normalization}

We next assess whether intensity normalization contributes to the observed performance differences. To this end, two new models were trained on Training Sets \#1 and \#2 using the normalization derived from Training Set \#3.

In both cases, performance improved compared to default normalization (Fig.~\ref{fig:curves}), with the largest gains observed in early low-contrast frames. This effect is also reflected in the phase-based analysis (Fig.~\ref{fig:boxplot}), where normalization increases median performance and reduces variability in the initial phase for both LA and LAA.

However, neither the model trained on Training Set \#1 nor that trained on Training Set \#2 with modified normalization matched the performance of Training Set \#3. This indicates that while nnUNet’s foreground-based normalization improves generalization, it does not fully compensate for the absence of low-contrast examples. For Training Set \#2, normalization also increases variability and slightly lowers median performance in later phases.

\begin{figure*}[t]
    \centering
    \includegraphics[width=0.94\linewidth]{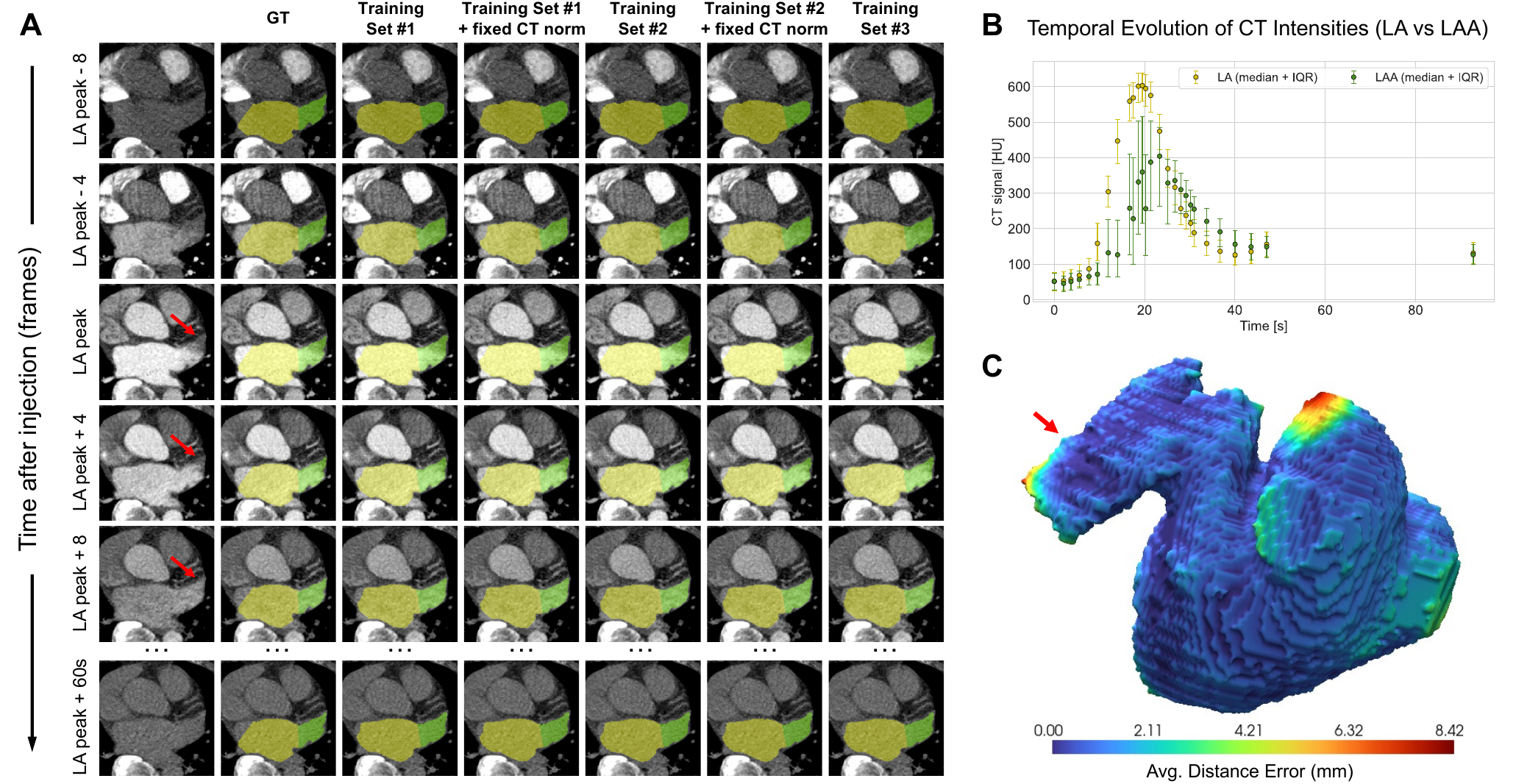}
    \caption{Qualitative and temporal analysis on a representative test subject. (A) Segmentation results from all models across six frames spanning different contrast phases, compared with ground truth (GT). Non-uniform filling of the LAA is indicated by a red arrow. (B) Temporal evolution of CT intensities (median and IQR) in the LA and LAA, showing greater variability and delayed peak enhancement in the LAA. (C) 3D distance rendering of the average segmentation error over the sequence for the model trained on Training Set \#3, computed as the distance from GT to prediction, projecting the errors over the GT mask. The average errors obtained are low, including in the late-rising area (red arrow).}
    \label{fig:qualitative}
\end{figure*}

Qualitative example (Fig.~\ref{fig:qualitative}) support these findings. Figure~\ref{fig:qualitative} shows segmentation results across representative frames from a test subject. In early low-contrast frames, Training Set \#3 provides more consistent delineation of the LAA, whereas reduced datasets show occasional under-segmentation or boundary leakage. Despite substantial intensity variability due to non-uniform LAA filling, all models remain robust in segmenting distal regions, supported by case stratification based on filling differences during training. Differences between models are reduced at peak and delayed phases, consistent with the convergence observed in Dice and HD95.

\section{Discussion}

In this study, we showed that temporal training-set design is a key factor for robust LA and LAA segmentation in dynamic contrast 4DCT. Training with all frames yielded the most consistent performance across the full sequence, particularly in early low-contrast phases. However, a physiologically selected subset of frames recovered most of this benefit while reducing temporal redundancy and potentially limiting supervision noise from propagated labels. The normalization ablation further indicates that nnUNet’s foreground-based CT normalization contributes to improved performance in low-contrast conditions, but does not fully compensate for the absence of low-contrast examples during training.

These findings suggest that the optimal training strategy should be guided by the downstream application. While full-sequence training is preferable when robustness across all phases is required, tasks focused on high-contrast frames may be effectively addressed using a small number of carefully selected, well-annotated images, thereby reducing the impact of propagated labeling errors. This is consistent with standard clinical practice, where typically only peak LA and delayed frames are acquired.

This is particularly relevant for applications such as quantification of blood flow in the LAA, where accurate segmentation during specific contrast phases may be sufficient to characterize patient-specific flow dynamics in patients with atrial fibrillation.

\section*{Acknowledgments}  
%
The authors acknowledge the support of the Ministerio de Ciencia e Innovación, Agencia Estatal de Investigación, under grants PDC2022-133865-I00 and PID2022-141493OB-I00 funded by MCIN/AEI/10.13039/ 501100011033, co-financed by European Regional Development Fund (ERDF/ EU), ‘A way of making Europe’ and the European Union ‘‘Next Generation EU’’/PRTR. This work was supported by the MAGERITCM project (grant TEC-2024/COM-44, Comunidad de Madrid, Spain). DMG's FPU PhD fellowship was funded by the Ministerio de Ciencia, Innovación y Universidades. 

\bibliography{refs}

\begin{correspondence}
David Montalvo-García [david.montalvo@upm.es]\\
María J. Ledesma-Carbayo [mj.ledesma@upm.es]\\
Av. Complutense 30, ETSIT UPM, 28040 Madrid
\end{correspondence}

\end{document}